# On Causal and Anticausal Learning

**Bernhard Schölkopf, Dominik Janzing, Jonas Peters, Eleni Sgouritsa, Kun Zhang**   FIRST.LAST@TUE.MPG.DE
Max Planck Institute for Intelligent Systems, Spemannstrasse, 72076 Tübingen, Germany

**Joris Mooij**   J.MOOIJ@CS.RU.NL
Institute for Computing and Information Sciences, Radboud University, Nijmegen, The Netherlands

## Abstract

We consider the problem of function estimation in the case where an underlying causal model can be inferred. This has implications for popular scenarios such as covariate shift, concept drift, transfer learning and semi-supervised learning. We argue that causal knowledge may facilitate some approaches for a given problem, and rule out others. In particular, we formulate a hypothesis for when semi-supervised learning can help, and corroborate it with empirical results.

## 1. Introduction

A large part of machine learning research aims at exploiting statistical associations or dependences between variables to make predictions about certain variables. This is useful especially in situations where we have sizable training sets, but no detailed model of the underlying data generating process. It has been argued that statistical associations are always due to underlying causal structures (Reichenbach, 1956). This suggests the question how machine learning could benefit from knowledge of these structures. The present paper addresses this question in the simplest possible setting, where the causal structure only consists of cause and effect, and there are no unobserved confounders. We argue that under certain assumptions, detailed below, there are asymmetries in joint distributions that have implications for statistical machine learning. We try to give a systematic outline of these implications. This paper does not prove theorems; rather, it aims at providing insight and drawing connections. It does not contain new experimental data, but a meta-analysis of performances reported by three other studies, focusing on an implication of causal structure for semi-supervised learning. We believe that the implications of causal structure for machine learning are conceptually intriguing, and we hope that they will raise interest for causal inference in the machine learning community.

An example illustrating the difference between the statistical and the causal point of view is the correlation between the frequency of storks and the human birth rate (Matthews, 2000). We may be able to train a good predictor of the birth rate which uses the frequency of storks (along with other features) as an input. However, if politicians asked us whether one could boost the birth rate by increasing the number of storks, we would have to tell them that this kind of *intervention* is not covered by the standard i.i.d. assumption of statistical learning. In practice, however, interventions can be relevant, distributions may shift over time, and we might want to combine data recorded under different conditions or from different but related regularities.

We briefly summarize some aspects of causal graphical models as pioneered by Pearl (2000); Spirtes et al. (1993). These are usually thought of as joint probability distributions over a set of variables $X_1, \ldots, X_n$, along with a directed acyclic graph with vertices $X_i$ and arrows indicating direct causal influences. The *causal Markov assumption* states that each vertex $X_i$ is independent of its non-descendants in the graph, given its parents. Crucially, this links causal semantics (which is important for predicting how a system reacts to interventions) to something that has empirically measurable consequences. Given observations from a joint distribution, it allows us to test conditional independence statements and thus infer (subject to a genericity assumption referred to as *faithfulness*) which causal models are consistent with an observed distribution. This will typically not lead us to a unique causal model though.

An alternative approach, referred to as a functional causal model (a.k.a. structural causal model or nonlinear structural equation model), starts with a set of jointly independent noise variables, one per vertex, and each vertex computes a deterministic function of its noise variable and its parents. These functions do not describe relations between observations only, but also how the system behaves under interventions: by changing the input of some of the functions, one can compute the effect of *setting* some variables to specific values. A functional model entails a joint distribution which along with the graph satisfies the causal Markov assumption (Pearl, 2000). Vice versa, each causal graphical model can be expressed as a functional causal model.

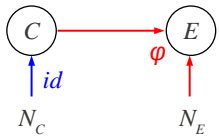

*Figure 1.* A simple functional causal model, where $C$ is the cause variable, $\varphi$ is a deterministic mechanism, and $E$ is the effect variable. $N_C$ is a noise variable influencing $C$ (without restricting generality, we can identify this with $C$), and $N_E$ influences $E$ via $E = \varphi(C, N_E)$. We assume that $N_C$ and $N_E$ are independent.

The functional point of view allows us to come up with assumptions on causal models that would be harder to conceive in a pure probabilistic view. Such assumptions (see below) allow us to distinguish between $X \rightarrow Y$ and $X \leftarrow Y$. This cannot be achieved by conditional independence testing, since no nontrivial conditional independences exist if we only have two variables.

The functional point of view thus opens up the possibility to infer the causal direction for input-output learning problems. Perhaps somewhat surprisingly, learning problems need not always predict effect from cause, and we will argue that the direction of the prediction has consequences for which tasks are easy and which tasks are hard.

**Notation.** We consider the causal structure shown in Fig. 1, with two observables, modeled by random variables. The variable $C$ stands for the cause and $E$ for the effect. We denote their distributions by $P(C)$ and $P(E)$ (overloading the notation $P$), and the domains by calligraphic symbols $\mathcal{C}$ and $\mathcal{E}$. The variable $X$ will always be the input and $Y$ the output (or prediction), but input and output can either be cause or effect — more below. For simplicity, we assume that their distributions have a joint density with respect to some product measure. We write the values of this density as $P(c, e)$ and the values of the marginal densities as $P(c)$ and $P(e)$, again keeping in mind that these three $P$ are different functions — we can always tell from the argument which function is meant. In some places, we will use conditional densities, always implicitly assuming that they exist.

The following assumptions are used throughout the paper.

**Causal sufficiency.** We assume that there are two independent noise variables $N_C$ and $N_E$, modeled as random variables with distributions $P(N_C)$ and $P(N_E)$.

The function $\varphi$ and the noise $N_E$ jointly determine $P(E|C)$ via $E = \varphi(C, N_E)$. We think of $P(E|C)$ as the *mechanism* transforming cause $C$ into effect $E$.[1]

---
[1] Note that we will use the term "mechanism" both for the

**Independence of mechanism and input.** We finally assume that the mechanism is "independent" of the distribution of the cause (i.e., independent of $P(C) = P(N_C)$, cf. Fig. 1), in the sense that $P(E|C)$ contains no information about $P(C)$ and vice versa; in particular, if $P(E|C)$ changes at some point in time, there is no reason to believe that $P(C)$ changes at the same time.[2]

This assumption has been used by Janzing & Schölkopf (2010), inspired by Lemeire & Dirkx (2007). It is plausible if we are dealing with a mechanism of nature that does not care what we feed into it. For instance, in the problem of predicting splicing patterns from genomic sequences, the basic splicing mechanism (driven by the ribosome) may be assumed evolutionarily stable and thus independent of the species (Schweikert et al., 2009), even though the genomic sequences and their statistical properties differ. Intuitively, if we learn a causal model of splicing, we could hope to be more robust with respect to changes of the input statistics.

The independence assumption introduces an asymmetry between cause and effect, since it will usually be violated in the backward direction, i.e., $P(E)$ and $P(C|E)$ are dependent because both inherit properties from $P(E|C)$ and $P(C)$ (Janzing & Schölkopf, 2010; Daniušis et al., 2010).

**Richness of functional causal models** It turns out that the two-variable functional causal model is so rich that the causal direction cannot be inferred. To understand the richness of the class intuitively, consider the simple case where the noise $N_E$ can take only a finite number of values, say $\{1, \ldots, v\}$. This noise could affect $\varphi$ for instance as follows: there is a set of functions $\{\varphi_n : n = 1, \ldots, v\}$, and the noise randomly switches one of them on at any point, i.e., $\varphi(c, n) = \varphi_n(c)$. The functions $\varphi_n$ could implement arbitrarily different mechanisms, and it would thus be hard to identify $\varphi$ from empirical data sampled from such a complex model. In view of this, it is surprising that conditional independence alone does allow us to do causal inference of practical significance, as implemented by the PC and FCI algorithms (Spirtes et al., 1993; Pearl, 2000). However, additional assumptions that prevent the noise switching construction can significantly facilitate the task of inferring causal graphs from data. Intuitively, such assumptions need to control the sensitivity of the mechanism $\varphi$ to the change in the noise $N_E$, and thus the complexity of $P(E|C)$.

**Additive noise models.** One such assumption is referred to as ANM, standing for *additive noise model* (Hoyer et al., 2009). This model assumes $\varphi(C, N_E) = \phi(C) + N_E$ for some function $\phi$:

$$E = \phi(C) + N_E, \qquad (1)$$

---
function $\varphi$ and for the conditional $P(E|C)$, but not for $P(C|E)$.

[2] This "independence" condition is closely related to the concept of exogeneity in economics (Pearl, 2000). Given two variables $C$ and $E$, we say $C$ is exogenous if $P(E|C)$ remains invariant to changes in the process that generates $C$.

and it has been shown that $\phi$ and $N_E$ can be inferred in the generic case, provided that $N_E$ has zero mean. This means that apart from some exceptions, such as the case where $\phi$ is linear and $N_E$ is Gaussian, a given joint distribution of two real-valued random variables $X$ and $Y$ can be fit by an `ANM` model in at most one direction (which we then consider the causal one). A similar statement has been shown for the `postnonlinear ANM` model (Zhang & Hyvärinen, 2009) $E = \psi(\phi(C) + N_E)$, where $\psi$ is an invertible function. In practice, an `ANM` model can be fit by regressing the effect on the cause while enforcing that the residual noise variable is independent of the cause (Mooij et al., 2009). If this is impossible, the model is incorrect (e.g., cause and effect are interchanged, the noise is not additive, or there are confounders; in the latter two cases the method cannot find the causal direction).

`ANM`s play an important role in this paper; first, the methods below will presuppose that we know what is cause and what is effect, and second, we will generalize `ANM` to handle the case where we have several models of the form (1) that share the same $\phi$. The following sections provide an overview of how causal direction affects various learning scenarios, partly relying on assumptions such as ANMs.

The comprehensive work of Storkey (2009) already describes the cases discussed in Sections 2.1.1 and 3.2.1, but not the other cases where further assumptions are needed. He also describes several scenarios where both $P(C)$ and $P(E|C)$ change, for instance if the data set is obtained by sample selection according to the value of a common effect of $C$ and $E$, or an effect of $E$, and the case where the data sets correspond to different values of a common cause of $C$ and $E$. Pearl & Bareinboim (2011) introduce a variable $S$ that labels different domains or data sets and explain how the way in which $S$ is causally linked to variables of interest is relevant for transferring causal or statistical statements across domains. Their notion of transportability employs conditional independences to express invariance of mechanisms, which is not general enough to include all the types of invariances we have in mind. For instance, the functions representing a causal mechanism could remain the same, while the unobserved noise terms may differ across data sets. Finally, we point out that an earlier version of the present work appeared as (Schölkopf et al., 2011).

## 2. Predicting Effect from Cause

Let us consider the case where we are trying to estimate a function $f : \mathcal{X} \to \mathcal{Y}$ or a conditional distribution $P(Y|X)$ in the causal direction, i.e., that $X$ is the cause and $Y$ the effect. Intuitively, this situation of *causal prediction* should be the 'easy' case since there exists a functional mechanism $\varphi$ which $f$ should try to mimic. We are interested in the question how robust the estimation is with respect to changes in the noise variables of the underlying model.

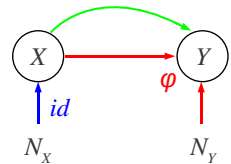

*Figure 2.* Predicting effect $Y$ from cause $X$.

### 2.1. Additional information about the input

#### 2.1.1. ROBUSTNESS W.R.T. INPUT CHANGES

**Given:** training points from $P(X, Y)$ and an additional set of inputs sampled from $P'(X)$, with $P(X) \neq P'(X)$.

**Goal:** estimate $P'(Y|X)$.

**Solution:** by independence of mechanism and input, there is no reason to assume that the observed change in $P(X)$ (i.e., in $P(N_X)$) entails a change in $P(Y|X)$, and we thus conclude $P'(Y|X) = P(Y|X)$. This scenario is referred to as *covariate shift* (Sugiyama & Kawanabe, 2012). The equation $P'(Y|X) = P(Y|X)$ should, however, not be mistaken as saying that the rule for predicting $Y$ from $X$ need not be adapted to the new input distribution $P(X)$. This is because prediction from finite data may favor simple functions that fit the data well in the region where $P(X)$ has high probability, but not where $P'(X)$ is high.

#### 2.1.2. SEMI-SUPERVISED LEARNING (SSL)

**Given:** training points sampled from $P(X, Y)$ and an additional set of inputs sampled from $P(X)$.

**Goal:** estimate $P(Y|X)$.

**Note:** by independence of the mechanism, $P(X)$ contains no information about $P(Y|X)$. A more accurate estimate of $P(X)$, as may be possible by the addition of the test inputs $P(X)$, does thus not influence an estimate of $P(Y|X)$, and SSL is pointless for the scenario in Figure 2.

### 2.2. Additional information about the output

#### 2.2.1. ROBUSTNESS W.R.T. OUTPUT CHANGES

**Given:** training points from $P(X, Y)$ and an additional set of outputs sampled from $P'(Y)$, with $P'(Y) \neq P(Y)$.

**Goal:** estimate $P'(Y|X)$.

**Assumption:** not clear

**Solution:** first we need to decide whether $P(X)$ or $P(Y|X)$ has changed, for some rough ideas see `Localizing distribution change` (Section 4). If $P(X)$ has changed, proceed as in Section 2.1.1. If $P(Y|X)$ has changed, we can estimate $P'(Y|X)$ via `Estimating causal conditionals` (Section 4). Here, additive noise is a sufficient assumption.

### 2.2.2. ADDITIONAL OUTPUTS

**Given:** training points sampled from $P(X,Y)$ and an additional set of outputs sampled from $P(Y)$.

**Goal:** estimate $P(Y|X)$.

**Assumption:** $P(X,Y)$ has an additive noise model from $X$ to $Y$ and $P(Y)$ has a unique decomposition as convolution of two distributions, say $P(Y) = Q * R$. This is, for instance, satisfied if the noise is Gaussian and $P(\phi(C))$ is indecomposable (i.e., it cannot be written as a non-trivial convolution of two distributions).

**Solution:** The additional outputs help because the decomposition tells us that either $P(N_Y) = Q$ or $P(N_Y) = R$. The additive noise model learned from the $x,y$-pairs will probably tell us which of the alternatives is true. Knowing $P(Y)$, learning $P(Y|X)$ reduces to learning $\phi$ from the $x,y$-pairs, which is a weaker problem than learning $P(Y|X)$ would be in general.

### 2.3. Additional information about input and output

#### 2.3.1. TRANSFER LEARNING (ONLY NOISE CHANGES)

**Given:** training points sampled from $P(X,Y)$ and an additional set of points sampled from $P'(X,Y)$, with $P'(X,Y) \neq P(X,Y)$.

**Goal:** estimate $P'(Y|X)$.

**Assumption:** Additive noise, where $\phi$ is invariant but the noises can change.

**Solution:** run `Conditional ANM` to output a single function, only enforcing independence of residuals separately for the two data sets (Section 4).

There is also a SSL variant of this scenario: Given a training set plus two unpaired sets from the two original marginals, the extra sets help to better estimate $P(X,Y)$ because we have argued in Section 2.2.2 that additional $y$-values sampled from $P(Y)$ already help.

The fact that causal directions matter for transferring knowledge from one data set to another one has previously been pointed out by Storkey (2009).

#### 2.3.2. CONCEPT DRIFT (ONLY FUNCTION CHANGES)

**Given:** training points sampled from $P(X,Y)$ and an additional set of points sampled from $P'(X,Y)$, with $P'(X,Y) \neq P(X,Y)$.

**Goal:** estimate $P'(Y|X)$.

**Assumption:** ANM with $N_X, N_Y$ invariant, but $\phi$ has changed.

**Solution:** Apply ANM to points sampled from $P'(X,Y)$ to obtain $\phi$. Then $P'(Y|X)$ is given by $P'(Y|X) =$

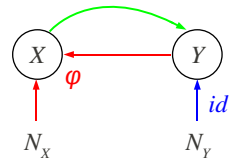

*Figure 3.* Predicting cause $Y$ from effect $X$.

$P_{N_Y}(Y - \phi(X))$, where the index $N_Y$ indicates the variable this distribution refers to.

## 3. Predicting Cause from Effect

We now turn to the opposite direction, where we consider the effect as input and we try to predict the value of the cause variable that led to it. This situation, that we refer to as *anticausal prediction*, may seem unnatural, but it is actually ubiquitous in machine learning. Consider, for instance, the task of predicting the class label of a handwritten digit from its image. The causal structure is as follows: a person intends to write the digit 7, say, and this intention causes a motor pattern producing an image of the digit 7 — in that sense the class label $Y$ causes the image $X$.

$P(X|Y)$ represents the causal mechanism that generates $X$ from $Y$, and it is independent from the distribution of the cause, $P(Y)$. On the other hand, $P(Y|X)$ is sensitive to the change of the distribution of $P(Y)$. Therefore, generally speaking, when estimating $P(Y|X)$, it would be better to model $P(X|Y)$ first and then construct $P(Y|X)$ using Bayes' rule $P(Y|X) = P(X|Y)P(Y)/P(X)$.

For a simple illustration, suppose $X = Y + N_X$, where both $Y$ and $N_X$ are independent from each other and uniformly distributed. Fig. 4 shows the scatter plot of $Y$ and $X$. The expectation $\mathbb{E}(X|Y)$ is linear in $Y$, and $P(X|Y)$ can be easily described. However, one can see that $\mathbb{E}(Y|X)$ is rather complex; it is nonlinear in $X$ and its shape depends heavily on the distribution of $P(Y)$.

### 3.1. Additional information about the input

#### 3.1.1. ROBUSTNESS W.R.T. INPUT CHANGES

**Given:** training points sampled from $P(X,Y)$ and an additional set of inputs from $P'(X)$, with $P'(X) \neq P(X)$.[3]

**Goal:** estimate $P'(Y|X)$.

**Assumption:** additive Gaussian noise with invertible function $\phi$ and indecomposable $P(\phi(Y))$ is sufficient. Other assumptions are also possible, but invertibility of the causal conditional $P(X|Y)$ is necessary in any case.

---

[3] A related scenario is that we do not have additional data from $P'(X)$, but we want to still use our knowledge of the causal direction to learn a model that is somewhat robust w.r.t. changes of $P(X)$ due to changes in either $P(Y)$ or $P(X|Y)$.

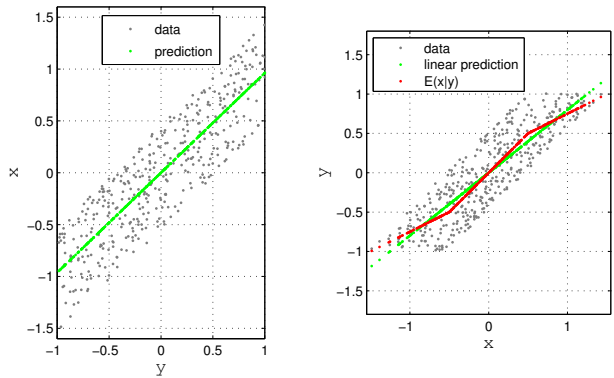

*Figure 4.* An illustration on the difference between predicting the effect and predicting the cause. Left: predicting effect $X$ from cause $Y$ under the causal relation $X = Y + N_X$, where $Y$ and $N_X$ are uniformly distributed. Right: predicting cause $Y$ from effect $X$.

**Solution:** We apply `Localizing distribution change` (Section 4) to decide if $P(Y)$ or $P(X|Y)$ has changed. In the first case, we can estimate $P'(Y)$ via `Inverting conditionals` (Section 4) if we assume that $P(X|Y)$ is an injective conditional.[4] From this we get $P'(X,Y)$, and then $P'(Y|X) = \frac{P'(X,Y)}{\int P'(X,Y)dY}$.

If, of the other hand, $P(X|Y)$ has changed, we can estimate $P'(X|Y)$ via `Estimating causal conditionals` (Section 4).

### 3.1.2. SEMI-SUPERVISED LEARNING

**Given:** training points sampled from $P(X,Y)$ and an additional set of inputs sampled from $P(X)$.

**Goal:** estimate $P(Y|X)$.

**Assumption:** various options, see below

**Note:** $P(X)$ and $P(Y|X)$ are not independent and thus contain information about each other. The additional inputs hence may allow a more accurate estimate of $P(X)$.

Known assumptions for SSL, as discussed by Chapelle et al. (2006), can indeed be viewed as linking properties of $P(X)$ to properties of $P(Y|X)$: for instance, the *cluster assumption* stipulates that points lying in the same cluster of $P(X)$ have the same $Y$; and the *low density separation assumption* states that the decision boundary of a classifier (i.e., the point where $P(Y|X)$ crosses 0.5) should lie in a region where $P(X)$ is small. The *semi-supervised smoothness assumption* says that the estimated function (which we may think of as the expectation of $P(Y|X)$) should be smooth in regions where $P(X)$ is large.

---

[4]Injectivity means that the input distribution can uniquely be computed from the output distribution, see Section 4.

### 3.2. Additional information about the output

#### 3.2.1. ROBUSTNESS W.R.T. OUTPUT CHANGES

**Given:** training points sampled from $P(X,Y)$ and an additional set of outputs sampled from $P'(Y)$, with $P'(Y) \neq P(Y)$. This scenario is also refered to as prior probability shift (Storkey, 2009).

**Goal:** estimate $P'(Y|X)$.

**Solution:** independence of mechanism implies $P'(X|Y) = P(X|Y)$, thus $P'(X,Y) = P(X|Y)P'(Y)$. From this, we compute $P'(Y|X) = \frac{P'(X|Y)P'(Y)}{\int P'(X,Y)dY}$.

### 3.3. Additional information about input and output

#### 3.3.1. ROBUSTNESS W.R.T. CHANGES OF INPUT AND OUTPUT NOISE (TRANSFER LEARNING)

**Given:** training points sampled from $P(X,Y)$ and an additional set of points sampled from $P'(X,Y)$, with $P'(X,Y) \neq P(X,Y)$.

**Goal:** estimate $P'(Y|X)$.

**Assumption:** additive noise where $\phi$ is invariant, but the noises can change.

**Solution:** analogous to Section 2.3.1, but use the model backwards in the end.

#### 3.3.2. CONCEPT DRIFT (CHANGE OF FUNCTION)

**Given:** training points sampled from $P(X,Y)$ and an additional set of points sampled from $P'(X,Y)$, with $P'(X,Y) \neq P(X,Y)$.

**Goal:** estimate $P'(Y|X)$.

**Assumption:** $N_X, N_Y$ invariant, but $\phi$ has changed to $\phi'$.

**Solution:** We can learn $\phi'$ from $P'(X,Y)$ and then estimate the entire distribution $P'(X,Y)$ using the estimates of $P(N_X)$ and $P(N_Y)$ obtained from observing those $x,y$ pairs that were sampled from $P(X,Y)$.

## 4. Modules

`Inverting conditionals` We can think of a conditional $P(Y|X)$ as a mechanism that transforms $P(X)$ into $P(Y)$. In some cases, we do not lose any information by this mechanism:

**Definition 1 (injective conditionals)** *a conditional distribution $P(Y|X)$ is called injective if there are no two distributions $P(X) \neq P'(X)$ such that*

$$\int P(y|x)P(x)dx = \int P(y|x)P'(x)dx.$$

**Example 1 (full rank stochastic matrix)** *Let $X, Y$ have*

finite range. Then $P(Y|X)$ is given by a stochastic matrix $M$ and is injective if and only if $M$ has full rank. Note that this is only possible if $|\mathcal{X}| \leq |\mathcal{Y}|$.

**Example 2 (post-nonlinear model)** *Let $X, Y$ be real-valued and let $Y = \psi(\phi(X) + N_Y)$ with $N_Y \perp\!\!\!\perp X$ be a post-nonlinear model where $\phi$ and $\psi$ are injective. Then the distribution of $Y$ uniquely determines the distribution of $\phi(X) + N_Y$ because $\psi$ is invertible. This in turn, uniquely determines the distribution of $\phi(X)$ provided that the convolution with $P(N_Y)$ is invertible. Since $\psi$ is invertible, this determines the distribution of $X$ uniquely.*

`Localizing distribution change` Given data points sampled from $P(C, E)$ and additional points from $P'(E) \neq P(E)$, we wish to decide whether $P(C)$ or $P(E|C)$ has changed. To show that appropriate assumptions render this problem solvable, we sketch some rough ideas. Let $E = \phi(C) + N_E$, with the same $\phi$ for both distributions $P(E, C)$ and $P'(E, C)$, but the distribution of the noise $N_E$ or the distribution of $C$ changes. Let $P(\phi(C))$ denote the distribution of $\phi(C)$.[5] Then the distributions of the effect are given by

$$P(E) = P(\phi(C)) * P(N_E),$$
$$P'(E) = P'(\phi(C)) * P'(N_E),$$

where either $P'(\phi(C)) = P(\phi(C))$ or $P'(N_E) = P(N_E)$. In the following situations, for instance, we can decide which of the cases is true:

1) If the Fourier transform of $P(E)$ contains zeros, then some of them correspond to zeros in the spectrum of $P(\phi(C))$, the others to zeros of the spectrum of $P(N_E)$. Then we may check which zeros still appear in $P'(E)$.

2) Suppose $P(\phi(C))$ and $P'(\phi(C))$ are indecomposable and $P(N_E)$ and $P'(N_E)$ are zero mean Gaussian; then the distribution $P(E) = P(\phi(C)) * P(N_E)$ uniquely determines $P(\phi(C))$ by deconvolving $P(E)$ with the Gaussian of maximal possible width that still yields a density.

`Estimating causal conditionals` Given $P'(E)$, estimate $P'(E|C)$ under the assumption that $P(C)$ remains constant. Assume that $P(E, C)$ and $P'(E, C)$ have been generated by the additive noise model $E = \phi(C) + N_E$, with the same $P(C)$ and $\phi$, while the distribution of $N_E$ has changed. We have

$$P(E) = P(\phi(C)) * P(N_E),$$
$$P'(E) = P(\phi(C)) * P'(N_E).$$

Hence, $P'(N_E)$ can be obtained by the deconvolution $P'(N_E) = P(\phi(C)) *^{-1} P'(E)$. This way, we can compute the new conditional $P'(E|C)$.

---

[5]Explicitly, it is derived from the distribution of $C$ by $P(\phi(C) \in A) = P(C \in \phi^{-1}(A))$.

`Conditional ANM` Given two data sets generated by $E = \phi(C) + N_E$ and $E' = \phi(C') + N'_E$, respectively. We modify the algorithm of Mooij et al. (2009) to obtain the shared function $\phi$, enforcing separate independence $C \perp\!\!\!\perp N_E$ and $C' \perp\!\!\!\perp N'_E$.

This can be interpreted as a generalized ANM model, enforcing *conditional* independence in $E|i = \phi(C|i) + N_E|i$, where $i \in \{1, 2\}$ is an index, and $C \perp\!\!\!\perp N_E \,|\, i$.

## 5. Empirical Results

An evaluation of all methods described is beyond the scope of this paper. We focus on assaying our main prediction regarding the difficulty of SSL; for a toy example applying `Conditional ANM` in transfer learning, see [1].

**Semi-supervised classification** We compare the performance of SSL algorithms with that of base classifiers using only labeled data. For many examples $X$ is vector-valued. We assign each dataset to one of three categories:

**1.** *Anticausal/Confounded:* (a) datasets in which at least one feature $X_i$ is an effect of the class $Y$ to be predicted (Anticausal) (includes also cyclic causal relations between $X_i$ and $Y$) and (b) datasets in which at least one feature $X_i$ has an unobserved common cause with the class $Y$ to be predicted (Confounded). In both (a) and (b) the mechanism $P(Y|X_i)$ can be dependent on $P(X_i)$. For these datasets, additional data from $P(X)$ may thus improve prediction.

**2.** *Causal:* datasets in which some features are causes of the class, and there is no feature which (a) is an effect of the class or (b) has a common cause with the class. If our assumption on independence of cause and mechanism holds, then SSL should be futile on these datasets.

**3.** *Unclear:* datasets which were difficult to be categorized to one of the aforementioned categories. Some of the reasons for that are incomplete documentation or lack of domain knowledge.

In practice, we count a dataset already as causal when we believe that the dependence between $X$ and $Y$ is *mainly* due to $X$ causing $Y$, although additional confounding effects may be possible.

We first analyze the results in the benchmark chapter of a book on SSL (Tables 21.11 and 21.13 of Chapelle et al. (2006)), for the case of 100 labeled training points. The chapter compares 11 SSL methods to the base classifiers 1-NN and SVM. In [1], we give details on our subjective categorization of the eight datasets used in the chapter.

In view of our hypothesis, it is encouraging to see (Figure 5) that SSL does not significantly improve the accuracy in the one causal dataset, but it helps in most of the anti-causal/confounded datasets. However, it is difficult to draw conclusions from this small collection of datasets; moreover, two additional issues may confound things: (1) the experiments were carried out in a *transductive* setting. In-ductive methods use unlabeled data to arrive at a classifier which is subsequently applied to an unknown test set; in contrast, transductive methods use the test inputs to make predictions. This could potentially allow performance improvements independent of whether a dataset is causal or anticausal; (2) the SSL methods used cover a broad range, and were not extensions of the base classifiers; moreover, the results for the SecStr dataset are based on a different set of methods than the rest of the benchmarks.

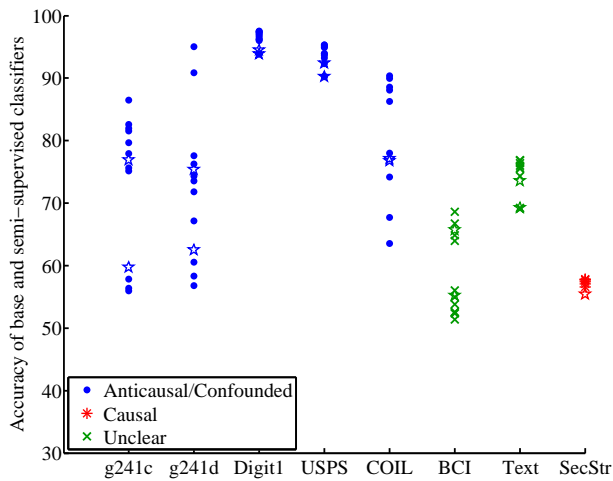

Figure 5. Accuracy of base classifiers (star shape) and different SSL methods on eight benchmark datasets.

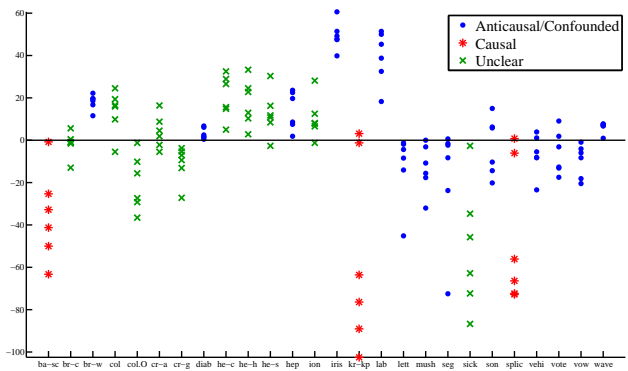

Figure 6. Plot of the relative decrease of error when using self-training, for six base classifiers on 26 UCI datasets. Here, relative decrease is defined as (error(base) − error(self-train)) / error(base). Self-training, a method for SSL, overall does not help for the causal datasets, but it does help for several of the anti-causal/confounded datasets.

We next consider 26 UCI datasets and six different base classifiers. The original results are from Tables III and IV in (Guo et al., 2010), and are presently re-analyzed in terms of the above dataset categories. The comprehensive results of Guo et al. (2010) allow us the luxury of (1) considering only self-training, which is an extension of supervised learning to unlabeled data in the sense that if the set of unlabeled data is empty, we recover the results of the base method (in this case, self-training would stop at the first iteration). This lets us compare an SSL method to its corresponding base algorithm. Moreover, (2) we included only the *inductive* methods considered by Guo et al. (2010), and not the *transductive* ones (cf. our discussion above).

The web page [1] describes our subjective categorization of the 26 UCI datasets into Anticausal/Confounded, Causal, or Unclear. e

In Figure 6, we observe that SSL does not significantly decrease the error rate in the three causal datasets, but it does increase the performance in several of the anticausal/confounded datasets. This is again consistent with our hypothesis that if mechanism and input are independent, SSL will not help for causal datasets.

**Semi-supervised regression (SSR)** Classification problems are often inherently asymmetric in that the inputs are continuous and the outputs categorical. It is worth reassuring that we obtain similar results in the case of regression. To this end, we consider the co-regularized least squares regression (co-RLSR) algorithm, compared to regular RLSR on 32 real-world data sets by Brefeld et al. (2006) (two of which are identical, so 31 data sets were considered). We categorized them into causal/anticausal/unclear, prior to the subsequent analysis.

We deemed seven of the data sets anticausal, i.e., the target variable can be considered as the cause of (some of) the predictors; Fig. 7 shows that SSR reduces the root mean square errors (RMSE) in all these cases. Nine of the remaining datasets can be considered causal, and Fig. 8 shows that there is usually little performance improvement for those. As Brefeld et al. (2006), we used the Wilcoxon signed rank test to assess whether SSR outperforms supervised regression, in the anticausal and causal cases. The null hypothesis is that the distribution of the difference between the RMSE produced by SSR and that by supervised regression is symmetric around 0 (i.e., that SSR does not help). On the anticausal datasets, the p-value is 0.0156, while it is 0.6523 on the causal datasets. Therefore, we reject the null hypothesis in the anticausal case at a 5% significance level, but not in the causal case.

## 6. Conclusion

If one is interested in predicting one variable from another one, it helps to know the causal structure underlying the variables. We give an overview of the implication of prediction in causal and anticausal directions, in particular formulating the hypothesis that under an independence assumption for causal mechanism and input, semi-supervised learning works better in anticausal or confounded problems than in causal problems. Our preliminary meta-analysis of results from the literature seems to support this claim.

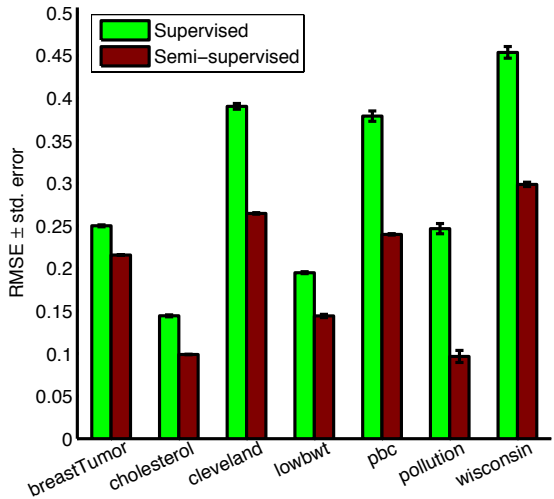

Figure 7. RMSE for Anticausal/Confounded datasets.

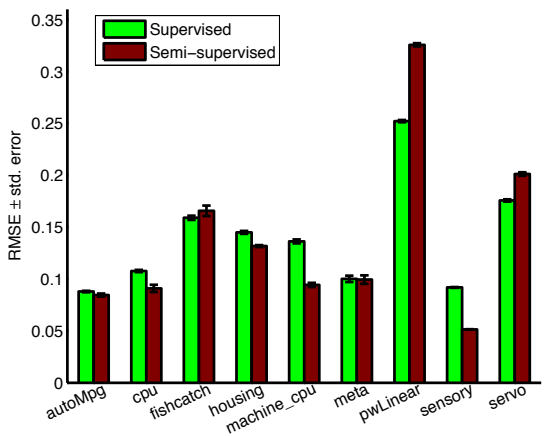

Figure 8. RMSE for Causal datasets.

**Acknowledgement** We thank Ulf Brefeld and Stefan Wrobel who kindly shared their detailed experimental results with us, allowing for our meta-analysis. We thank Bob Williamson, Vladimir Vapnik, and Jakob Zscheischler for helpful discussions.